# Transfer Learning as an Enhancement for Reconfiguration Management of Cyber-Physical Production Systems

Benjamin Maschler [a,*], Timo Müller [a], Andreas Löcklin [a], Michael Weyrich [a]

[a] *University of Stuttgart, Institute of Industrial Automation and Software Engineering, Pfaffenwaldring 47, 70569 Stuttgart, Germany*

* Corresponding author. Tel.: +49 711 685 67295; Fax: +49 711 685 67302. *E-mail address:* benjamin.maschler@ias.uni-stuttgart.de

**Abstract**

Reconfiguration demand is increasing due to frequent requirement changes for manufacturing systems. Recent approaches aim at investigating feasible configuration alternatives from which they select the optimal one. This relies on processes whose behavior is not reliant on e.g. the production sequence. However, when machine learning is used, components' behavior depends on the process' specifics, requiring additional concepts to successfully conduct reconfiguration management. Therefore, we propose the enhancement of the comprehensive reconfiguration management with transfer learning. This provides the ability to assess the machine learning dependent behavior of the different CPPS configurations with reduced effort and further assists the recommissioning of the chosen one. A real cyber-physical production system from the discrete manufacturing domain is utilized to demonstrate the aforementioned proposal.

## 1. Introduction

In recent years, machine learning algorithms' practical utilization in cyber-physical production systems (CPPS) has been a major area of research [1, 2]. Many of these aim at providing new functionalities as an addition to conventional control software, e.g. in order to detect anomalies [3], provide autonomy [4] or predict faults [5]. However, it remains challenging to integrate those into productive industrial systems, at least partially because of the latters' frequent changes calling for frequent re-trainings of the algorithms.

One driver of such long-term system dynamics are the trends towards shorter innovation and product life cycles [6] as well as the development towards mass individualization which lead to frequent changes of production requirements. Because these changes in requirements render the production system goals unpredictable during the design phase, adaptions during the operational phase, i.e. reconfigurations, are gaining in importance [7].

However, in order to plan reconfiguration measures and assess their consequences for systems utilizing machine learning, i.e. systems whose control software potentially behaves differently after those changes are implemented, one needs to consider a reconfiguration's impact on the algorithms as well. This further increases the aforementioned need for frequent re-trainings of algorithms utilized in live CPPS – a challenge that could be met by transferring knowledge between different variants of the same algorithm, reducing the computational effort and the amount of training data needed [8].

*Objective*: In this article, case studies from the domain of industrial manufacturing underlining the potentials of reconfiguration management enhanced with transfer learning capabilities are presented and qualitatively analyzed.

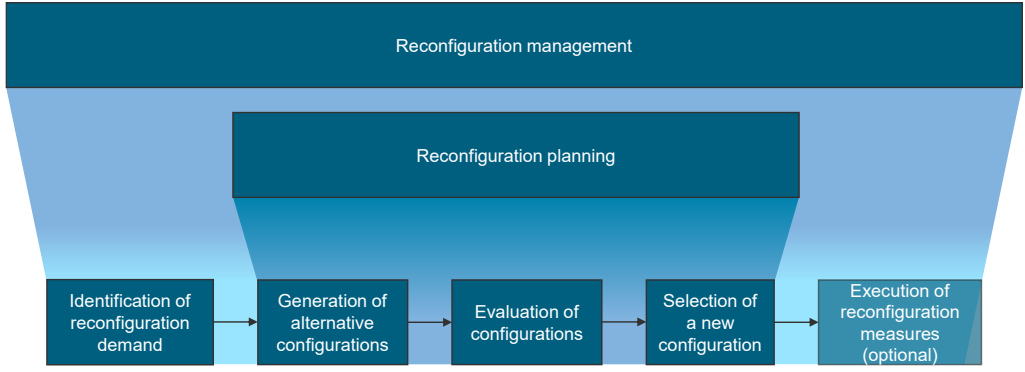

Fig. 1. Scope of reconfiguration activities

*Structure*: In chapter 2, related work on the topics of reconfiguration management and transfer learning is introduced. From there, a proposal is derived in chapter 3. Chapter 4 subsequently presents case study scenarios as well as the cyber-physical production system used therein. Finally, a conclusion and an outlook are given in chapter 5.

## 2. Related Work

### 2.1. Reconfiguration Management

The scope of the reconfiguration activities (see Fig. 1) extends beyond the mere execution of reconfiguration measures. Therefore, the term *reconfiguration management* is introduced in [7] and [9] and can be summarized "to span the identification of reconfiguration demand, the generation of alternative configurations, the evaluation of configurations, the selection of a new configuration and, as an optional extension, the execution of reconfiguration measures" [10]. Thereby, the *generation of alternative configurations*, the *evaluation of configurations* and the *selection of a new configuration* can be aggregated to describe the process of reconfiguration planning.

Regarding the *identification of reconfiguration demand*, Table I introduces diverse reconfiguration triggers which are categorized as *internal* or *external triggers*. However, this list does not claim to be comprehensive, because new triggers can and will emerge due to the continued volatility of requirements in industrial automation.

Except for the *execution of reconfiguration measures*, a comprehensive reconfiguration management is carried out in the cyber layer, utilizing digital models of the CPPS and running them in simulations, possibly in recourse to a pre-existing digital twin [11]. The complete process can be summarized in the following manner:

Whenever one of the mentioned triggers occurs, a comparison of the target production with the current CPPS configuration has to be performed in order to identify a possible reconfiguration demand.

As soon as a reconfiguration demand has been identified, the *generation of alternative configurations* is initiated. In this step, different CPPS configurations that are capable of conducting the target production sequence are discovered. In order make sure to include the most suitable reconfiguration, simulation-based optimization as proposed in [10] can be carried out.

For the *evaluation of configurations* of the (optimized) alternatives, defined metrics to provide a quantitative basis for the decision-making should be applied. These metrics may contain non-functional criteria such as time, cost or energy, both in terms of reconfiguration and operation, i.e. post-reconfiguration production, efforts.

Subsequently, the *selection of a new configuration* is conducted. The evaluated configuration alternatives of the CPPS are compared based on the results of the previous step.

Finally, the *execution of reconfiguration measures,* i.e. the implementation of the hard- and software changes necessary to convert the pre-reconfiguration CPPS into a CPPS resembling the previously selected new configuration, is carried out. Classically, this is achieved in a manual approach, but it represents an optional extension for the reconfiguration management if done (partially) automatically.

### 2.2. Transfer Learning

In machine learning terminology, transfer learning refers to the transfer of knowledge and skills from previously learned tasks to new tasks in order to improve a learning system's performance on the latter [12]. Special methods are required to achieve such a transfer as merely continuing the training on a dataset representing the new task is usually not sufficient. This is caused by different marginal distributions between the respective datasets, requiring different predictive functions [8, 13, 14].

Table 1: Internal and external reconfiguration triggers

| Internal triggers | External triggers |
| --- | --- |
| A system component encounters a defect. | A new product or product variant is requested. |
| Maintenance measures become necessary. | A changed production volume is requested. |
| The applied configuration does not achieve the expected metrics. | An increased flexibility of the production system is requested. |
| | An improvement of the quality characteristics of the production system is requested. |
| | The legal situation changes (e.g. COVID-19 demands for more personal space). |
| | The environmental requirements change. |
| | The availability of required production material changes. |

Whereas, traditionally, a lot of this topic's literature focusses on the application of transfer learning on visual recognition (esp. in the medical domain), natural language processing or communication tasks [14, 15], there has recently been increasing interest from the industrial domain as well [8, 16]. Here, transfer learning is utilized to overcome two problems hindering a more universal deployment of deep learning:

- Due to only very low numbers of identical industrial machinery or systems, high standards of data protection and low levels of cooperation between different enterprises, sufficiently large and diverse datasets for successful training are hard to acquire [13, 17].
- Due to increasing demand for frequent reconfigurations [18], changing processes and dynamic environments, quickly outdating datasets once acquired only provide short-term representations of the problem space necessitating continuous data collection and algorithm retraining [16, 19].

Transfer learning mitigates those challenges by enabling algorithms to train not only on datasets characterizing the task at hand, but on related, e.g. simulated, ones as well. Furthermore, it allows algorithms to adapt to changing tasks without requiring retraining from scratch.

Some examples from the industrial domain illustrate its potential: In [20], the authors created an algorithm capable of single-shot object recognition learning on dynamic tasksets. In [21], an anomaly detection algorithm tolerating changing input data dimensionality is presented. In [13], the authors pre-trained a quality management algorithm on simulated data before fine-tuning it on the actual dataset collected from the real process.

This transfer of knowledge between different lifecycle phases or different modes of data, i.e. simulated and real data, as used in the last example is termed *cross-phase industrial transfer learning*. It represents one of four base use cases of industrial transfer learning according to [8] (see Fig. 2). So far, only very few publications address this use case. This might be due to a perceived lack of beneficial applications or a lack of high-quality datasets combining data collected from simulated and real assets.

## 3. Proposed approach

Whenever CPPS with machine learning functionalities are reconfigured, the reconfiguration measures will potentially have an impact on the machine learning component's behavior, because after reconfiguration they will have to operate under potentially altered conditions. Therefore, any reconfiguration measures' impact on such CPPS's machine learning components needs to be considered during the generation and evaluation of alternative configurations.

Depending on the specific implementation of such alternatives' generation, more or less variants of a CPPS are to be created. However, it is safe to assume that even a small number of variants will pose a great challenge if their respective machine learning algorithms have to be trained from scratch, i.e. random initialization.

Therefore, the hypotheses that transfer learning would lead to a reduction of the computational burden and training data demand for the retraining of machine learning algorithms caused by a CPPS's reconfiguration arises. Thus, we propose to combine the two concepts of reconfiguration management and transfer learning.

Such a combination could benefit the initial creation of a simulation model based on the pre-reconfiguration CPPS (Real2Sim transfer), the generation of alternative simulation models (Sim2Sim transfer) as well as the rollout of the post-reconfiguration CPPS (Sim2Real transfer). These opportunities for the application of transfer learning with respect to reconfiguration activities of a CPPS utilizing machine learning are depicted in Fig. 3.

## 4. Case Study

In this chapter, the potential of the enhancement of reconfiguration management with transfer learning for CPPS utilizing machine learning is discussed using the example of an actual production system.

### 4.1. Cyber-physical Production System

To investigate the possibilities of reconfigurable manufacturing systems, a cyber-physical production system based on the modules from Festo's cyber-physical factory platform is used. In this Cyber-physical Production Lab at the Institute of Industrial Automation and Software Engineering (IAS-CPP-Lab), five workstations, an automatic pallet warehouse and up to four automated guided vehicles (AGV) for transportation purposes are installed (see Fig. 4). The AGVs are based on the Festo Robotino 3 Premium platform and can automatically dock to any workstation. Each stationary module

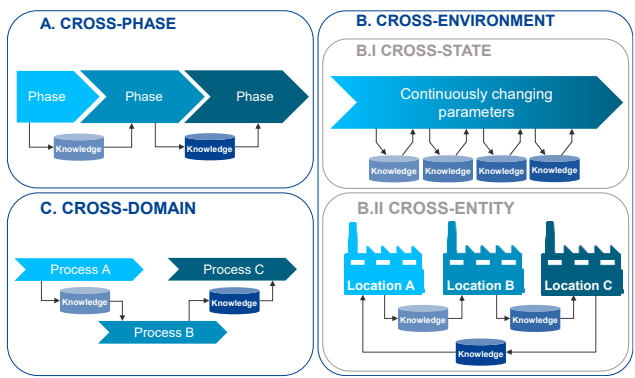

Fig. 2. Industrial transfer learning base use cases according to [7]

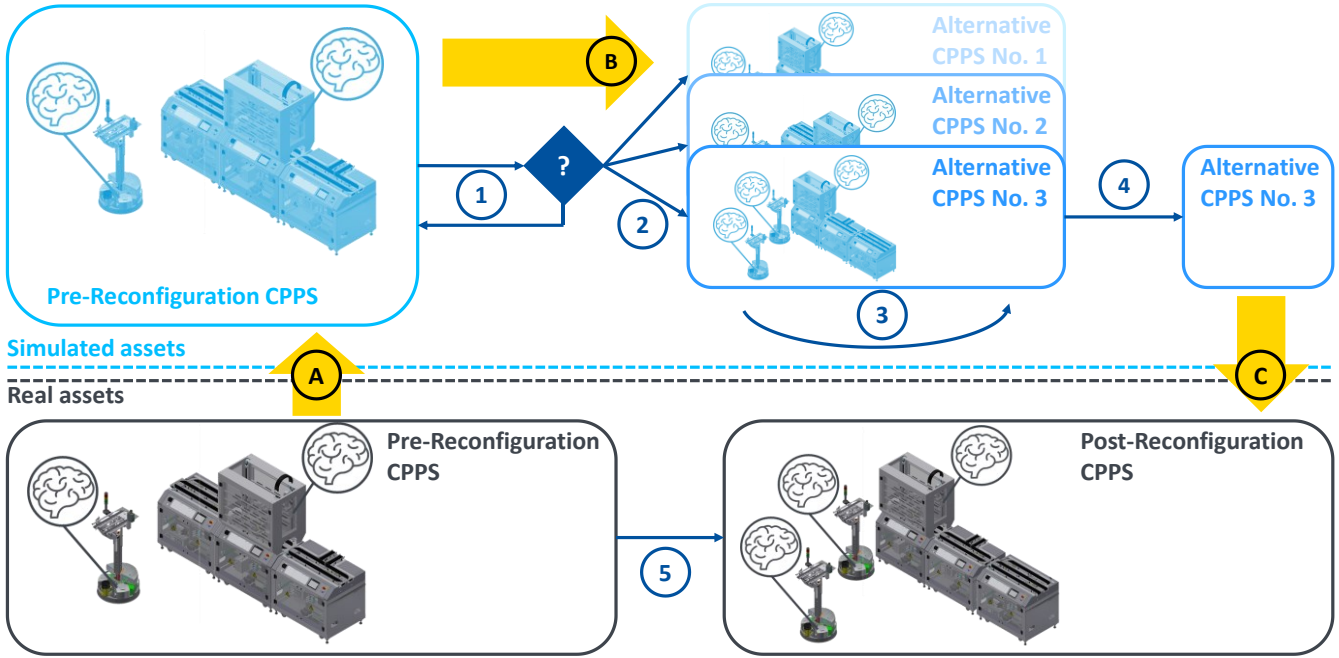

Fig. 3. Opportunities for transfer learning in the reconfiguration of a CPPS utilizing machine learning (1: Identification of reconfiguration demand; 2: Generation of alternative configurations; 3: Evaluation of alternative configurations; 4: Selection of new configuration; 5: Execution of reconfiguration measures; A: Real2Sim transfer; B: Sim2Sim transfer; C: Sim2Real transfer)

is controlled by a separate programmable logic controller (PLC). The AGVs feature an onboard Linux PC. As a superordinate control, the manufacturing execution system (MES) Festo MES4 is responsible for coordinating the assembly of a printed circuit board with fuses and a housing. Any intralogistics tasks can be performed automatically using the Festo Fleet-Manager that serves as a link between MES and AGVs.

The following functionalities are implemented: The automatic pallet warehouse provides pallets that can hold parts such as unassembled printed circuit boards, fuses or front and rear housings. A robotic arm places those pallets on respectively retrieves pallets from workpiece carriers. Workpiece carriers are transferred among workstations and between workstations and AGVs via conveyor belts. Photoelectric barriers as well as RFID and inductive sensors are used to control the material flow. PLC and MES use Profinet for data transmission. The fleet manager and AGVs communicate via a separate WLAN.

At the beginning of a discrete manufacturing process, individual parts are requested from the warehouse. Assembly then takes place at the workstations. AGVs automatically dock at the workstations to deliver or pick up (semi-)finished products. Due to the high flexibility of AGV-based intralogistics, the layout of the realized modular production process can be changed at will. Depending on the order situation, up to four AGVs are used to handle transport tasks. Instead of AGVs, human workers can also transport workpiece carriers manually. RFID-based tracking of the workpiece carriers is used to detect incorrect deliveries.

At the IAS-CPP-Lab, the AGVs utilize machine learning to optimize their routing within the dynamically changing environment consisting of stationary modules, moving AGVs and human workers. Furthermore, machine learning can be utilized to improve material flow and the production sequences themselves.

### 4.2. Enhancement 1: Real2Sim Transfer

Since a comprehensive reconfiguration management (except for the execution of reconfiguration measures) is carried out within the cyber layer, first, simulation models of the CPPS need to be created – unless a fully synchronized digital twin already exists [22]. If the CPPS utilizes machine learning, naturally, the machine learning algorithms must be re-created as well. However, due to a shift of the marginal distributions of the domain space (see chapter 2.2.), a mere copy will not suffice – the simulated CPPS or its surroundings will differ from the real ones, causing the machine learning algorithms to behave differently as well, resulting in the so-called *reality gap* [13, 23].

One approach to tackle this challenge is to retrain the machine learning algorithms inside the simulation using simulated data. However, this is computational very costly as this data needs to be collected, first, whereupon the complete training needs to be carried out.

An easier alternative would be a transfer of the pre-trained, machine learning algorithms from reality into the simulation, the so-called *Real2Sim* transfer learning (see Fig. 3, letter A). It potentially requires less simulated data and a less intensive training process – speeding up the implementation of reconfiguration management for any CPPS utilizing machine learning.

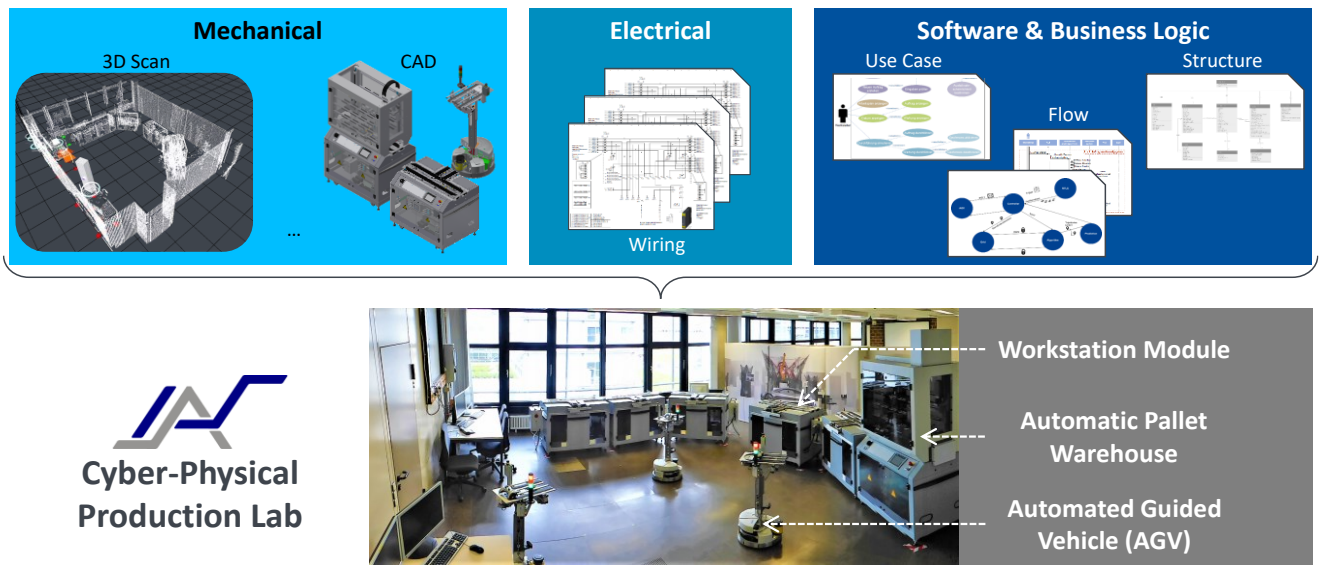

Fig. 4. Cyber-physical Production Lab at the Institute of Industrial Automation and Software Engineering

**Example**: To implement a reconfiguration management for the IAS-CPP-Lab, simulation models need to be created which must include simulation versions of all machine learning algorithms present. However, the simulated data will not be exactly the same as the real data, because the simulation will differ however minutely from reality. This necessitates an adaption of the simulation's machine learning algorithms, in this example optimizing the AGVs' routing, in order to mimic the non-simulated algorithms' behavior despite using simulated data instead of real data – i.e. it needs to address the reality gap. To retrain the algorithms in a conventional way requires a lot of simulated data and a complete training effort, whereas a retraining using transfer learning can be conducted using less data and computing resources.

### 4.3. Enhancement 2: Sim2Sim Transfer

During the second step of a comprehensive reconfiguration management, the generation of alternatives (see Fig. 3, number 2), different variants of the CPPS are created in simulation. If this CPPS utilizes machine learning, naturally, there must be different variants of the machine learning algorithms created as well. However, due to a shift of the marginal distributions of the domain space (see chapter 2.2.), a mere copy will not suffice – the whole point of the different CPPS's variants being their difference, the machine learning algorithms' variants will have to be different as well.

One approach to their creation is to retrain the machine learning algorithms' variants inside the simulation using simulated data. However, this is computational very costly as this data needs to be accumulated for each CPPS variant, first, whereupon the complete training needs to be carried out for each algorithm variant.

An easier alternative would be a transfer of the pre-trained, machine learning algorithms from the pre-reconfiguration simulation model into the different variants, the so-called *Sim2Sim* transfer learning (see Fig. 3, letter B). It potentially requires far less simulated data and a less intensive training process – speeding up the whole reconfiguration management procedure or enabling the exploration of a broader reconfiguration solution space within the same time for any CPPS utilizing machine learning.

**Example**: If twenty different alternative configurations, in this example involving different layouts or different numbers of AGVs, would be created for the IAS-CPP-Lab, twenty different machine learning algorithms have to be adapted to these. To retrain the algorithms in a conventional way requires a lot of simulated data and a complete training effort – twenty times. Retraining using transfer learning can be conducted using less data and computing resources. This way, even small gains through transfer learning have a great impact on the overall retraining effort.

### 4.4. Enhancement 3: Sim2Real Transfer

During the final step of a comprehensive reconfiguration management, the execution of reconfiguration measures (see Fig. 3, number 5), the real CPPS is altered in order to resemble the chosen alternative configuration. If this CPPS utilizes machine learning, naturally, the machine learning algorithms must be altered as well. Because of the shift of the marginal distributions of the domain space (see chapter 2.2.), a mere copy of the algorithms from simulation will not suffice due to the aforementioned reality gap (see chapter 4.2).

One approach to tackle this challenge is to retrain the machine learning algorithms on the reconfigured CPPS using newly collected data. However, this is very costly as the CPPS can oftentimes not be productively used while the data is accumulated and the complete training still needs to be carried out afterwards.

An easier alternative would be a transfer of the pre-trained, machine learning algorithms from the simulation of the chosen alternative configuration into reality, the so-called *Sim2Real* transfer learning (see Fig. 3, letter C). It potentially requires less newly collected data and a less intensive training process – speeding up the recommissioning of any reconfigured CPPS utilizing machine learning.

**Example**: After the reconfiguration measures have been executed, i.e. after the IAS-CPP-Lab has been reconfigured, the machine learning algorithms, in this example optimizing the AGVs' routing, need to be adapted, too. Conventionally, this requires the system to be run for a long time to acquire training data, whereafter the algorithms are trained anew. With transfer learning, the already similar machine learning algorithms from the chosen simulation model can be retrained requiring less training data, i.e. less time during which the system is run to collect such data, and less computational effort, speeding up the process of getting the reconfigured system fully operational.

## 5. Conclusion

The presented scenarios highlight potential benefits that arise from enhancing a comprehensive reconfiguration management with cross-phase transfer learning for CPPS utilizing machine learning algorithms. To make the scenarios more concrete, the cyber-physical production lab at the Institute of Industrial Automation and Software Engineering is used as an example.

Our main findings are:
- Real2Sim transfer learning can reduce the effort of creating simulation models involving machine learning capabilities. Those models are necessary for reconfiguration management. The effort reduction allows for a wider utilization of this concept.
- Sim2Sim transfer learning can greatly reduce the computational effort of creating alternative configurations involving machine learning components during reconfiguration management. This allows for a cheaper and/or more thorough exploration of possible configurations.
- Sim2Real transfer learning can reduce the effort of training machine learning algorithms on the reconfigured system, i.e. after reconfiguration measures have been executed. This reduces the time needed to get the system fully operational again.

Evaluations of the quantitative extend of these benefits are still ongoing and will be included in later publications. However, although the very first implementations have been carried out, there still is a substantial need for further research regarding a greater variety of reference implementations, benchmark comparisons and more real-life applications.